\def\year{2022}\relax
\documentclass[letterpaper]{article} %
\usepackage{aaai22}  %
\usepackage{times}  %
\usepackage{helvet}  %
\usepackage{courier}  %
\usepackage[hyphens]{url}  %
\usepackage{graphicx} %
\usepackage{natbib}  %
\usepackage{caption} %
\DeclareCaptionStyle{ruled}{labelfont=normalfont,labelsep=colon,strut=off} %
\usepackage{algorithm}
\usepackage{algorithmic}

\usepackage{graphicx}
\usepackage{multirow}
\usepackage{booktabs}
\usepackage{color,colortbl}
\usepackage{subcaption}
\usepackage{enumitem}
\usepackage{adjustbox}
\usepackage[utf8]{inputenc}

\usepackage{placeins}

\usepackage{amsmath,amsfonts,bm}

\def\eqref#1{equation~\ref{#1}}

\def\1{\bm{1}}

\def\vp{{\bm{p}}}

\def\vt{{\bm{t}}}

\def\mW{{\bm{W}}}

\DeclareMathAlphabet{\mathsfit}{\encodingdefault}{\sfdefault}{m}{sl}
\SetMathAlphabet{\mathsfit}{bold}{\encodingdefault}{\sfdefault}{bx}{n}

\usepackage{newfloat}
\usepackage{listings}
\floatstyle{ruled}
\newfloat{listing}{tb}{lst}{}
\floatname{listing}{Listing}
\nocopyright
\usepackage{xcolor}

\definecolor{LightCyan}{rgb}{0.88,1,1}

\title{BROS: A Pre-trained Language Model Focusing on Text and Layout\\for Better Key Information Extraction from Documents}
\author {
    Teakgyu Hong\textsuperscript{\rm 1},
    Donghyun Kim\textsuperscript{\rm 1},
    Mingi Ji\textsuperscript{\rm 2},
    Wonseok Hwang\textsuperscript{\rm 3},
    Daehyun Nam\textsuperscript{\rm 4},
    Sungrae Park\textsuperscript{\rm 4}
}
\affiliations {
    \textsuperscript{\rm 1}NAVER CLOVA, \textsuperscript{\rm 2}KAIST, \textsuperscript{\rm 3}LBox, \textsuperscript{\rm 4}Upstage AI Research, Upstage AI \\
    teakgyu.hong@navercorp.com, dong.hyun@navercorp.com, qwertgfdcvb@kaist.ac.kr,\\ wonseok.hwang@lbox.kr, daehyun.nam@upstage.ai, sungrae.park@upstage.ai
}

\begin{document}

\maketitle

\begin{abstract}

Key information extraction (KIE) from document images requires understanding the contextual and spatial semantics of texts in two-dimensional (2D) space.
Many recent studies try to solve the task by developing pre-trained language models focusing on combining visual features from document images with texts and their layout.
On the other hand, this paper tackles the problem by going back to the basic: effective combination of text and layout. 
Specifically, we propose a pre-trained language model, named \textit{BROS (BERT Relying On Spatiality)}, that encodes relative positions of texts in 2D space and learns from unlabeled documents with area-masking strategy.
With this optimized training scheme for understanding texts in 2D space, BROS shows comparable or better performance compared to previous methods on four KIE benchmarks (FUNSD, SROIE$^*$, CORD, and SciTSR) without relying on visual features.
This paper also reveals two real-world challenges in KIE tasks--(1) minimizing the error from incorrect text ordering and (2) efficient learning from fewer downstream examples--and demonstrates the superiority of BROS over previous methods.
\emph{Code is available at \url{https://github.com/clovaai/bros}.}

\end{abstract}

\section{Introduction}

Automatic key information extraction (KIE) from industrial documents is an essential task in robotic process automation (RPA).
Extracting an ordered item list from receipts~\cite{park2019cord}, prices and taxes from invoices~\cite{liu2019graph}, and paired key-values from form-like documents~\cite{jaume2019funsd} are representative examples.
Since the task requires understanding texts in various layouts, the combination of multiple technical components from both computer vision and natural language processing is required.

\begin{figure}
    \centering
    \includegraphics[width=\linewidth]{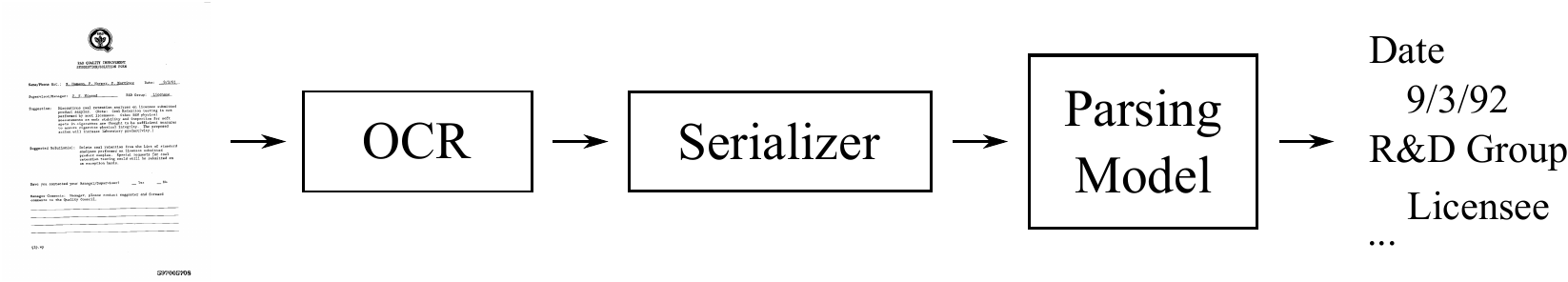}
    \caption{Schematic illustrations of document KIE pipeline.}
    \label{fig:doc_kie_pipeline}
\end{figure}

Figure~\ref{fig:doc_kie_pipeline} describes a schematic illustration of pipeline for the document KIE tasks~\cite{POST,BERTgrid}.
First, given a document image, optical character recognition (OCR) detects the texts in the image and recognizes the content to generate a set of text blocks.
Next, a serializer identifies a reading order of text blocks distributed in 2D image space and converts them into text sequence in 1D text space to apply NLP technology which is developed for 1D text sequence.
The most basic form of serializer is to arrange text blocks in a top-to-bottom and left-to-right way~\cite{clausner2013significance}.
Finally, from the serialized text blocks, key information is extracted via the parsing model.

\begin{table}[t]
    \vspace{-0.5em}
    \tabcolsep = 3pt
    \centering
    \begin{tabular}{lccccc}
    \toprule
        Model & Img & \# Params & (O) & (P) & (F)\\
        \midrule
        LayoutLM$_{\text{BASE}}$ &  & 113M & 78.66 & 33.89 & 62.50\\
        LayoutLMv2$_{\text{BASE}}$ & $\circ$ & 200M & 82.76 & 40.77 & 69.92\\
        BROS$_{\text{BASE}}$ &  & \textbf{110M} & \textbf{83.05} & \textbf{76.94} & \textbf{72.60}\\
        \midrule
        LayoutLM$_{\text{LARGE}}$ &  & 343M & 78.95 & 33.11 & 61.00\\
        LayoutLMv2$_{\text{LARGE}}$ & $\circ$ & 426M & 84.20 & 62.53 & 72.12\\
        BROS$_{\text{LARGE}}$ &  & \textbf{340M} & \textbf{84.52} & \textbf{79.42} & \textbf{74.42}\\
        \bottomrule
    \end{tabular}
    \caption{Performance comparison of pre-trained language models on (O)riginal, (P)ermuted, and (F)ew training samples FUNSD KIE tasks. In (F), 10 samples are used. 
    }
    \label{tbl:teaser}
    \vspace{-1em}
\end{table}

In the first step, an off-the-shelf OCR tool is often employed as industrial documents consist of relatively clean characters compared to general scene text images.
On the other hand, there are no such off-the-shelf tools for the serializer
even though it is often non-trivial to determine the proper reading order of text blocks~\cite{ECCV-ORDER,LayoutReader}.
Representative examples are documents including multi-columns or multiple tables.
This absence of the general-purpose serializer implies the careful design of parsing module is necessary to robustly handle documents with complex layouts where the reading orders can be often incomplete.

In the early studies of KIE, accurate document parsing greatly depends on the order of text blocks.
Once the serializer identifies an order, the set of the text blocks are converted into a text sequence and processed via a language model such as BERT~\cite{BERT} to identify key information~\cite{POST,BERTgrid}.
The linguistic understanding of the pre-trained language model leads to superior performance than rule-based extractions.
However, the conversion of texts in 2D space into a text sequence in 1D space leads to the loss of layout information that is critical in KIE tasks.

To avoid the loss of layout information, a new type of language model, LayoutLM~\cite{LayoutLM} expands a 1D positional encoding of BERT to 2D and is trained over a large corpus of industrial documents to understand spatial dependencies between text blocks.
Its fine-tuning has shown breakthrough performances on multiple KIE tasks and becomes a strong baseline.
After the rise of LayoutLM, several studies try to develop pre-trained language models by combining additional visual features~\cite{LayoutLMv2,TILT,SelfDoc,DocFormer,StrucTexT} (e.g. image patches identified by an object detection) and show further performance improvements.
However, the extensions using visual features require additional computational costs and they still demand more effective combinations of texts and their spatial information.

In this paper, we introduce a new pre-trained language model, named BROS, by re-focusing on the combinations of texts and their spatial information without relying on visual features.
Specifically, we propose an effective spatial encoding method by utilizing relative positions between text blocks, while most of previous works employ absolute 2D positions.
Additionally, we introduce a novel self-supervision method, named area-masked language model, that hides texts in an area of a document and supervises the masked texts.
With these two approaches for encoding of spatial information, BROS shows superior or comparable performances compared to previous methods using additional visual features.

Aside from improving KIE performances, BROS also addresses two important real-world challenges in KIE tasks: minimizing dependency on the order of text blocks and learning from a few training examples of downstream tasks.
The first challenge indicates the robustness on the serialization followed by the OCR process in Figure~\ref{fig:doc_kie_pipeline}.
In real scenario, document images are usually irregular (i.e. rotated or distorted documents) and the serializer might fail to identify a proper order of text blocks.
In addition, when serialization fails, the performance of sequence tagging approaches (e.g. BIO tagging), which most previous works employ, drops dramatically.  
To circumvent the difficulty, we apply SPADE~\cite{SPADE} decoder that extracts key text blocks without any order information to the pre-trained models and evaluates them on the new benchmarks where the order of text blocks are permuted.
As a result, BROS shows better robustness on the serializers compared to LayoutLM~\cite{LayoutLM} and LayoutLMv2~\cite{LayoutLMv2}.

The second challenge is related to the required number of labeled examples to understand the target key contents.
Since a single KIE example consists of hundreds of text blocks that should be categorized, the annotation is expensive.
Most public benchmarks consist of less than 1,000 samples, even though the target documents contain hundreds of layouts and diverse contexts.
In this paper, we analyze KIE performances over the number of training examples and compare the pre-trained models.
As a result, BROS performs better on FUNSD KIE tasks, and also BROS only with 20$\sim$30\% of FUNSD examples achieves better performance than LayoutLM with 100\% of them.
Summarized results for these experiments are shown in Table~\ref{tbl:teaser}.

Our contributions can be summarized as follows:
\begin{itemize}
    \item We propose an effective spatial layout encoding method by accounting for relative positions of text blocks.
    \item We also propose a novel area-masking self-supervision strategy that reflects 2D natures of text blocks. 
    \item The proposed model achieves comparable performance to the state-of-the-art without relying on visual features.
    \item We compare existing pre-trained models on permuted KIE datasets that lost the orders of text blocks.
    \item We compare the fine-tuning efficiency of various pre-trained models under a data-scarce environment.
\end{itemize}

\section{Related Work}

\subsection{Pre-trained Language Models for 2D Text Blocks}

Unlike the pre-trained models for conventional NLP tasks, such as BERT~\cite{BERT}, LayoutLM~\cite{LayoutLM} is first proposed to jointly model interaction between text and layout information for the document KIE task.
It encodes the absolute position of text blocks with axis-wise embedding tables and learns a token-level masked language model that hides tokens randomly and estimates the origins.
After the publication of LayoutLM, several pre-trained models have been tried to additionally integrate visual features, such as visual feature maps from raw images~\cite{LayoutLMv2,DocFormer}, image patches identified by an object detection module~\cite{SelfDoc}, and visual representations of text blocks~\cite{TILT,StrucTexT}. 
Although the extensions imposing multi-modalities of visual and textual features provide additional performance gains in KIE tasks, they spend additional computations to process raw document images.
Additionally, an effective combination of text and layout is still required as the major component of the multi-modalities.

Aside from incorporating visual features, StructuralLM~\cite{StructuralLM} utilizes cell information, a group of ordered text blocks, and shows promising performance improvements.
However, the local orders of text blocks might not be available depending on the KIE tasks and the OCR engines.
Therefore, this paper focuses on the original granularity of text blocks identified by OCR engines and improves the combination of text and layout by an effective spatial encoding method and an area-based pre-training strategy.

\begin{figure*}
    \centering
    \includegraphics[width=0.98\linewidth]{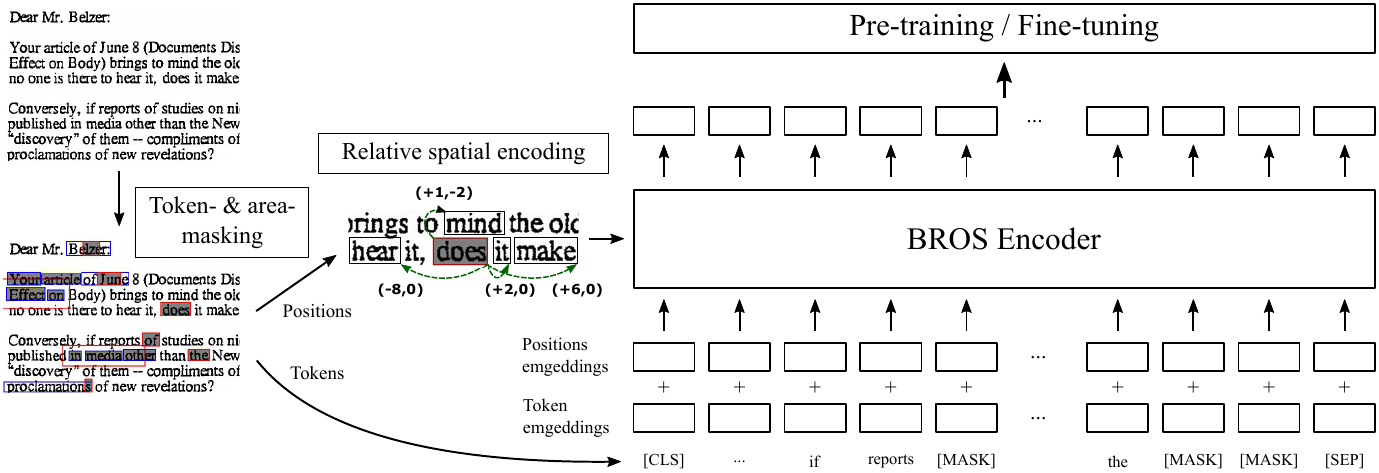}
    \caption{An overview of BROS. The tokens in the document image are masked through token- and area-masking strategy. The position difference between text blocks is encoded directly to the attention mechanism in Transformer. The output token representations are used in both pre-training and fine-tuning.}
    \label{fig:overview}
    \vspace{-1em}
\end{figure*}

\subsection{Parsers for Document Key Information Extraction}

BIO tagger, which is a representative parser for entity extraction from the text sequences, extracts key information by identifying spans with the beginning (B) and inside (I) points. %
Though BIO tagger has been used as a conventional method, it has two limitations for applying to document KIE.
One is that the correct order of text blocks is required for extracting key information when post-processing each classified token class (i.e. B- and I- classes).
For example, if the text blocks are not ordered properly, such as ``recognition, optical, character'', the correct answer can not be made.
The other is that it cannot solve the tasks that require the relationship between tokens as it performs token-level classification.

To overcome the above two limitations, we adopt a graph-based parser, SPADE~\cite{SPADE} decoder, that creates a directed relation graph of tokens to represent key entities and their relationships for KIE tasks. 
For example, SPADE can determine ``optical'' as a starting word and ``recognition'' as the next word.
By directly identifying relations between tokens, SPADE enables a description of all key information of KIE tasks regardless of the order of text blocks. 
In this paper, we apply the SPADE decoder for entity linking tasks of KIE benchmarks and also for all tasks lost perfect order information of text blocks.
Specifically, we slightly modify the SPADE decoder for better application with the pre-trained models.
The details can be found in the Appendix.

\section{BERT Relying on Spatiality (BROS)}

The main structure of BROS follows LayoutLM~\cite{LayoutLM}, but there are two critical advances: (1) a use of spatial encoding metric that describes spatial relations between text blocks and (2) a use of 2D pre-training objective designed for text blocks on 2D space.
Figure~\ref{fig:overview} shows a visual description of BROS for document KIE tasks.

\subsection{Encoding Spatial Information into BERT}

\begin{figure}[t]
    \captionsetup[subfigure]{aboveskip=1pt,belowskip=1pt}
    \centering
    \begin{subfigure}{0.8\linewidth}
        \centering
        \includegraphics[width=0.94\linewidth]{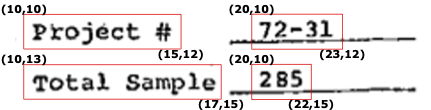}
        \caption{Encodes absolute spatial information.}
    \end{subfigure}
    \begin{subfigure}{0.8\linewidth}
        \centering
        \includegraphics[width=0.9\linewidth]{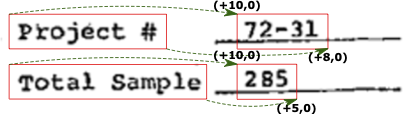}
        \caption{Encodes relative spatial information.}
    \end{subfigure}
    \caption{Comparison between absolute and relative positions. ``Project \#'' and ``Total Sample'' have their paired values, ``72-31'' and ``285'', respectively. In (a), the paired text blocks have different modalities based their absolute positions. On the other hand, in (b), they can hold co-modality to represent positions of their semantically coupled text blocks.}
    \label{fig:abs_vs_rel_pos}
    \vspace{-0.5em}
\end{figure}

\begin{figure}[tb]
    \captionsetup[subfigure]{aboveskip=5pt,belowskip=5pt}
    \centering
    \begin{subfigure}{.45\textwidth}%
        \centering
        \includegraphics[width=0.9\linewidth]{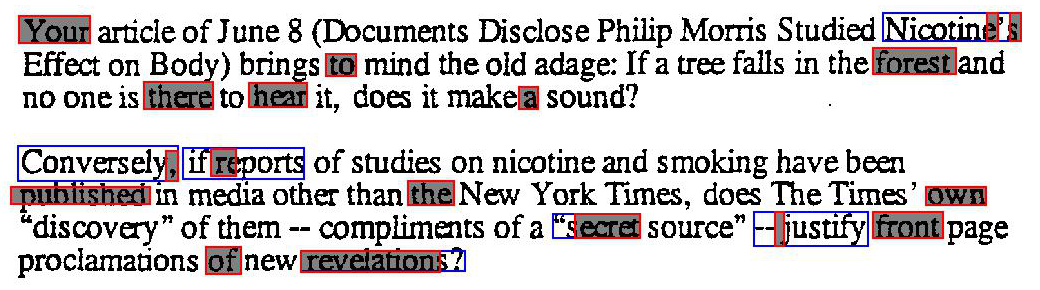}
        \caption{Random \textit{token} selection ({\color{red}{red}}) and \textit{token} masking ({\color{gray}{gray}})}
    \end{subfigure}
    \begin{subfigure}{.45\textwidth}%
        \centering
        \includegraphics[width=0.9\linewidth]{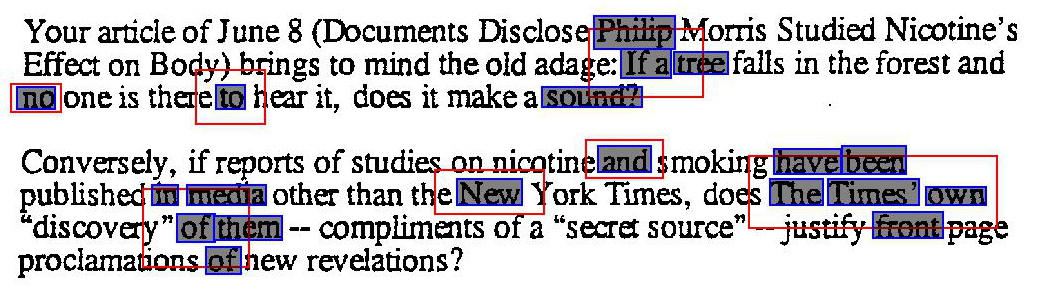}
        \caption{Random \textit{area} selection ({\color{red}{red}}) and \textit{block} masking ({\color{gray}{gray}})}
    \end{subfigure}
    \caption{Illustrations of two masking strategies.
    The {\color{blue}{blue}} boxes represent text blocks including masked tokens. In both figures,  15\% of tokens are masked.
    }
    \label{fig:masking}
    \vspace{-1em}
\end{figure}

The way to encode spatial information of text blocks decides how text blocks be aware of their spatial relations.
LayoutLM~\cite{LayoutLM} simply encodes absolute x- and y-axis positions to each text blocks but the specific-point encoding is not robust on the minor position changes of text blocks.
Instead, BROS employs relative positions between text blocks to explicitly encode spatial relations.
As shown in Figure~\ref{fig:abs_vs_rel_pos}, relative positions provides co-modality of spatial relations between text blocks regardless of their absolute position. 
This property can make the model better recognize entities which have similar key-value structures.

For formal description, we use $\vp = (x, y)$ to denote a point on 2D space and a bounding box of a text block consists of four vertices, such as $\vp^{\text{tl}}$, $\vp^{\text{tr}}$, $\vp^{\text{br}}$, and $\vp^{\text{bl}}$, that indicate top-left, top-right, bottom-right, and bottom-left points, respectively.
BROS first normalizes all the 2D points of the text blocks using the size of the image.
Then, BROS calculates relative positions of the vertices from the same vertices of the other bounding boxes of text blocks and applies sinusoidal functions as $\bar{\vp}_{i,j} = [\mathbf{f}^{\text{sinu}}(x_i - x_j); \mathbf{f}^{\text{sinu}}(y_i - y_j)]$.
Here, $\mathbf{f}^{\text{sinu}}: \mathbb{R} \rightarrow \mathbb{R}^{D^\text{s}}$ indicates a sinusoidal function, which is used in \citet{Transformer}, $D^\text{s}$ is the dimensions of sinusoid embedding, and the semicolon (;) indicates concatenation.
Through the calculations, the relative positions of $j^{\text{th}}$ bounding box based on the $i^{\text{th}}$ bounding box are represented with the four vectors, such as $\bar{\vp}_{i,j}^{\text{tl}}$, $\bar{\vp}_{i,j}^{\text{tr}}$, $\bar{\vp}_{i,j}^{\text{br}}$, and $\bar{\vp}_{i,j}^{\text{bl}}$.
Finally, BROS combines the four relative positions by applying a linear transformation, 
\begin{align}
    \overline{\bm{bb}}_{i,j} & = \mW^{\text{tl}}\bar{\vp}_{i,j}^{\text{tl}}+\mW^{\text{tr}}\bar{\vp}_{i,j}^{\text{tr}}+ \mW^{\text{br}}\bar{\vp}_{i,j}^{\text{br}}+\mW^{\text{bl}}\bar{\vp}_{i,j}^{\text{bl}}.
\end{align}
where $\mW^{\text{tl}}$, $\mW^{\text{tr}}$, $\mW^{\text{br}}$, $\mW^{\text{bl}} \in \mathbb{R}^{(H/A) \times 2D^\text{s}}$ are linear transition matrices, $H$ is a hidden size of BERT, and $A$ is the number of self-attention heads.

In the process of identifying the relative positional vector, $\overline{\bm{bb}}_{i,j}$, we carefully apply two components: the sinusoidal function, $\mathbf{f}^{\text{sinu}}$, and the shared embeddings to multiple heads of the attention module. %
First, the sinusoidal function can encode continuous distances more naturally than using a grid embedding that split a real-valued space into finite number of grids.
Second, the multi-head attention modules in Transformer share the same relative positional embeddings to impose the common spatial relationships between text blocks to multiple semantic features identified by the multiple heads.

BROS directly encodes the spatial relations to the contextualization of text blocks.
In detail, it calculates an attention logit combining both semantic and spatial features as follows; 
\begin{equation}
    \label{eqn:pe}
    a_{i, j}^h = ( \mW^{\text{q}}_h \vt_i )^\top (\mW^{\text{k}}_h \vt_j) + ( \mW^{\text{q}}_h \vt_i )^\top\overline{\bm{bb}}_{i,j},
\end{equation}
where $\vt_i$ and $\vt_j$ are context representations for $i^{\text{th}}$ and $j^{\text{th}}$ tokens and both $\mW^{\text{q}}_h$ and $\mW^{\text{k}}_h$ are linear transition matrices for $h^{\text{th}}$ head.
The former is the same as the original attention mechanism in Transformer~\citep{Transformer}.
The latter, motivated by \citet{TransformerXL}, considers the relative spatial information of the target text block when the source context and location are given.
As we mentioned above, we have shared relative spatial embedding across all of the different attention heads for imposing the common spatial relationships.

Compared to the spatial-aware attention in \citet{LayoutLMv2}, which utilizes axis-specific positional difference of text blocks as an attention bias, it has two major differences.
First, our method couples the relative embeddings with the semantic information of tokens for better conjugation between texts and their spatial relations. 
Second, when calculating the relative spatial information between two text blocks, we consider all four vertices of the block.
By doing this, our encoding can incorporate not only relative distance but also relative shape and size which play important roles in distinguishing key and value in a document.
We compare our relative encoding method and that of LayoutLMv2's in the ablation study.

\subsection{Area-masked Language Model}

Pre-training diverse layouts from unlabeled documents is a key factor for document KIE tasks. 
BROS utilizes two pre-training objectives: one is a token-masked LM (TMLM) used in BERT and the other is a novel area-masked LM (AMLM) introduced in this paper.
The area-masked LM, inspired by SpanBERT~\citep{SpanBERT}, captures consecutive text blocks based on a 2D area in a document.

TMLM randomly masks tokens while keeping their spatial information, and then the model predicts the masked tokens with the clues of spatial information and the other un-masked tokens.
The process is identical to MLM of BERT and Masked Visual-Language Model (MVLM) of LayoutLM.
Figure~\ref{fig:masking} (a) shows how TMLM masks tokens in a document.
Since tokens in a text block can be masked partially, their estimation can be conducted by referring to other tokens in the same block or text blocks near the masked token. 

AMLM masks all text blocks allocated in a randomly chosen area.
It can be interpreted as a span masking for text blocks in 2D space.
Specifically, AMLM consists of the following four steps: (1) randomly selects a text block, (2) identifies an area by expanding the region of the text block, (3) determines text blocks allocated in the area, and (4) masks all tokens of the text blocks and predicts them.
At the second step, the degree of expansion is identified by sampling a value from an exponential distribution with a hyper-parameter, $\lambda$.
The rationale behind using exponential distribution is to convert the geometric distribution used in SpanBERT for a discrete domain into a distribution for a continuous domain.
Thus, we set $\lambda=-\text{ln}(1-p)$ where $p=0.2$ used in SpanBERT.
Also, we truncated exponential distribution with 1 to prevent an infinity value covering all spaces of the document.
It should be noted that the masking area is expanded from a randomly selected text block since the area should be related to the text sizes and locations to represent text spans in 2D space.
Figure~\ref{fig:masking} compares token- and area-masking on text blocks.
Because AMLM hides spatially close tokens together, their estimation requires more clues from text blocks far from the estimation targets. 

Finally, BROS combines two masked LMs, TMLM and AMLM, to stimulate the model to learn both individual and consolidated token representations.
It first masks 15\% of tokens for AMLM and then masks 15\% of tokens on the left text blocks for TMLM.
Similar to BERT~\citep{BERT}, the masked tokens are replaced by \texttt{[MASK]} token for 80\%, random token for 10\%, and original token for the rest 10\%.

\section{Key Information Extraction Tasks}

\begin{figure}[t]
    \captionsetup[subfigure]{aboveskip=4pt,belowskip=4pt}
    \centering
    \begin{subfigure}{.45\textwidth}%
        \centering
        \includegraphics[width=0.9\linewidth]{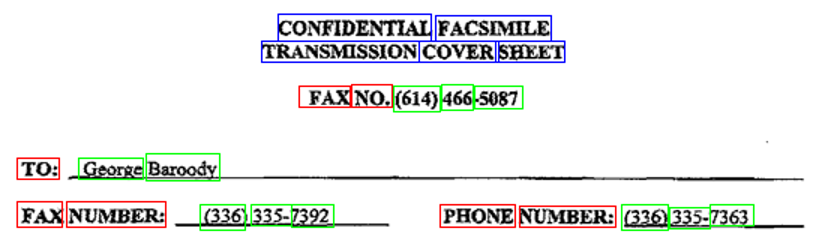}
        \caption{An example of FUNSD EE task.}
    \end{subfigure}
    \begin{subfigure}{.45\textwidth}%
        \centering
        \includegraphics[width=0.9\linewidth]{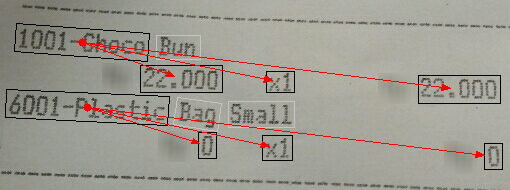}
        \caption{An example of CORD EL task.}
    \end{subfigure}
    \caption{Examples of EE and EL tasks. In (a), the colored blocks represent key entities. In (b), the red arrows show the hierarchical relationships between the entities.}
    \label{fig:ee_and_el_tasks}
\end{figure}

We solve two categories of KIE tasks, entity extraction (EE) and entity linking (EL). 
The EE task identifies sequences of text blocks that represent desired target texts.
Figure~\ref{fig:ee_and_el_tasks} (a) is an example of the EE task: identifying header, question, and answer entities in the form-like document.
The EL task connects key entities through their hierarchical or semantic relations.
Figure~\ref{fig:ee_and_el_tasks} (b) is an example of the EL task: grouping menu entities, such as its name, unit price, amount, and price.
Table~\ref{tbl:datasets} lists four KIE benchmark datasets: FUNSD~\citep{jaume2019funsd}, SROIE$^*$~\citep{huang2019icdar2019}, CORD~\citep{park2019cord}, and SciTSR~\citep{chi2019complicated}.
Details of the benchmarks can be found in Appendix.

\begin{table}[t]
    \centering
    \begin{adjustbox}{width=23.00em}
    \begin{tabular}{l|l|l|l}
        Dataset  & Types & Tasks & \# Images \\
        \toprule
        FUNSD    & Forms    & EE, EL & Train 149, Test 50 \\
        SROIE$^{*\dagger}$& Receipts   & EE     & Train 526, Test 100 \\
        CORD     & Receipts   & EE, EL & Train 800, Val 100, Test 100 \\
        SciTSR   & Tables    & EL     & Train 12,000, Test 3,000 \\
        \bottomrule    
        \multicolumn{4}{l}{\small
        $^\dagger$ modified version of SROIE. See details in Appendix.
        }
    \end{tabular}
    \end{adjustbox}
    \caption{Tasks and the number of images for each dataset.}
    \label{tbl:datasets}
    \vspace{-0.5em}
\end{table}

Although these four datasets provide testbeds for the EE and EL tasks, they represent the subset of real problems as the order information of text blocks is given.
FUNSD provides the orders of text blocks related to target classes in both training and testing examples.
In SROIE$^*$, CORD, and SciTSR, the text blocks are serialized in reading orders. 
To reflect the real scenario that does not contain perfect order information of text blocks, we remove the order information of KIE benchmarks by randomly permuting the order of text blocks.
We denote the permuted datasets as p-FUNSD, p-SROIE$^*$, p-CORD, and p-SciTSR.

\begin{table*}[t!]

\centering

\begin{adjustbox}{width=46em}
\begin{tabular}{l|l|rrr|r}
 & & \multicolumn{3}{c|}{FUNSD EE} & \\
Model & \multicolumn{1}{c|}{Modality} & \multicolumn{1}{c}{Precision} & \multicolumn{1}{c}{Recall} & \multicolumn{1}{c|}{F1} & \# Params \\
\toprule
BERT$_{\text{BASE}}$ \citep{LayoutLM}   & Text & 54.69 & 67.10 & 60.26 & 110M \\
LayoutLM$_{\text{BASE}}$ \citep{LayoutLM} & Text + Layout & 75.97 & 81.55 & 78.66 & 113M \\
DocFormer$_{\text{BASE}}$ \citep{DocFormer} & Text + Layout & 77.63 & 83.69 & 80.54 & 149M \\
\midrule
BROS$_{\text{BASE}}$ (Ours) & Text + Layout & \textbf{81.16}{\tiny $\pm$0.33} &  \textbf{85.02}{\tiny $\pm$0.32} &  \textbf{83.05}{\tiny $\pm$0.26} & 110M \\
\midrule
LayoutLM$_{\text{BASE}}$ \citep{LayoutLM} & Text + Layout + \textit{Image}* & 76.77 & 81.95 & 79.27 & 160M \\
LayoutLMv2$_{\text{BASE}}$ \citep{LayoutLMv2} & Text + Layout + \textit{Image}* & 80.29 & 85.39 & 82.76 & 200M \\
DocFormer$_{\text{BASE}}$ \citep{DocFormer} & Text + Layout + \textit{Image}* & 80.76 & \underline{86.09} & 83.34 & 183M \\
SelfDoc \citep{SelfDoc} & Text + Layout + \textit{Image}* & - & - & \underline{83.36} & 137M \\
StrucTexT \citep{StrucTexT} & Text + Layout + \textit{Image}* & \underline{85.68} & 80.97 & 83.09 & 107M$^\dagger$ \\
\midrule
\midrule
BERT$_{\text{LARGE}}$ \citep{LayoutLM}    & Text & 61.13 & 70.85 & 65.63 & 340M \\
LayoutLM$_{\text{LARGE}}$ \citep{LayoutLMv2} & Text + Layout & 75.96 & 82.19 & 78.95 & 343M \\
\midrule
BROS$_{\text{LARGE}}$ (Ours) & Text + Layout & \textbf{82.81}{\tiny $\pm$0.35} & \textbf{86.31}{\tiny $\pm$0.28} & \textbf{84.52}{\tiny $\pm$0.30} & 340M \\
\midrule
LayoutLMv2$_{\text{LARGE}}$ \citep{LayoutLMv2} & Text + Layout + \textit{Image}* & 83.24 & 85.19 & 84.20 & 426M \\
StructuralLM$_{\text{LARGE}}$ \citep{StructuralLM} & Text + Layout + \textit{Cell}* & \underline{83.52} & \underline{86.81} & \underline{85.14} & 355M \\
\bottomrule
\multicolumn{6}{l}{\small $^\dagger$ 
The number of parameters except for ResNet-FPN processing document images.
}
\end{tabular}
\end{adjustbox}
\caption{Performance comparison on the FUNSD EE task. \textbf{Bold} indicates the best performance among models using only text and layout, and \underline{underline} represents the best one. \textit{Image}* and \textit{Cell}* denote additional visual and hierarchical information, respectively. Our methods are repeatedly evaluated five times and the values of other methods are the reported scores.}
\label{tbl:tbl_funsd}
\vspace{-0.5em}
\end{table*}

\section{Experiments}

\subsection{Experiment Settings}

For pre-training, IIT-CDIP Test Collection 1.0\footnote{https://ir.nist.gov/cdip/} \citep{Lewis2006building}, which consists of approximately 11M document images, is used but 400K of RVL-CDIP dataset\footnote{https://www.cs.cmu.edu/~aharley/rvl-cdip/} \citep{harley2015evaluation} are excluded following LayoutLM.
To obtain text blocks from document images, CLOVA OCR API\footnote{https://clova.ai/ocr} was applied.
We observed no difference in performance depending on the OCR engine; LayoutLM trained in our experimental setting shows comparable performances to the published LayoutLM.
The sanity check can be found in Appendix.

The main Transformer structure of BROS is the same as BERT.
We set the hidden size, the number of self-attention heads, the feed-forward/filter size, and the number of Transformer layers of BROS$_{\text{BASE}}$ to 768, 12, 3072, and 12, respectively and those of BROS$_{\text{LARGE}}$ to 1024, 24, 4096, and 24, respectively.
The dimensions of sinusoid embedding $D^s$ is set to 24 for BROS$_{\text{BASE}}$ and 32 for BROS$_{\text{LARGE}}$.

BROS is trained by using AdamW optimizer~\citep{AdamW} with a learning rate of 5e-5 with linear decay.
The batch size is set to 64.
During pre-training, the first 10\% of the total epochs are used for a warm-up learning rate.
We initialized weights of BROS with those of BERT and trained it for 5 epochs on the IIT-CDIP dataset using 8 NVIDIA Tesla V100 32GB GPUs.

During fine-tuning, the learning rate is set to 5e-5.
The batch size is set to 16 for all tasks.
The number of training epochs or steps is as follows: 100 epochs for FUNSD, 1K steps for SROIE$^*$ and CORD, and 7.5 epochs for SciTSR.

\subsection{Experiment Results}

To evaluate the performance of the model, we first conduct experiments using the given order of text blocks in the dataset.
Then, we verify the robustness of the model against two important challenges in the KIE tasks, which are the dependency about the order of text blocks and learning from a few training examples.

Over our experiments, we report the scores of LayoutLM and LayoutLMv2 using the models published by the authors\footnote{https://github.com/microsoft/unilm} and denote them as LayoutLM$^*$ and LayoutLMv2$^*$.
We report the mean (and optionally the standard deviation) of the results using the 5 different random seeds.

\subsubsection{With the Order Information of Text Blocks}

Table~\ref{tbl:tbl_funsd} summarizes the results for the FUNSD EE task reported by previous approaches. %
When comparing models only using text and layout, BROS shows remarkable performance improvements by 2.51 (80.54 $\rightarrow$ 83.05) for the BASE models and 5.57 (78.95 $\rightarrow$ 84.52) for the LARGE models from the previous best.  
Interestingly, BROS provides better or similar performances compared to the multi-modal models incorporating additional visual (\textit{Image$^*$}) or hierarchical (\textit{Cell$^*$}) information. 
In other words, although BROS does not require extra computations and parameters to process additional features, BROS can achieves better or comparable performances.

\begin{table}[t]
\tabcolsep = 3pt
\centering
\begin{adjustbox}{width=23.5em}
\begin{tabular}{l|rrr|rrr}
 & \multicolumn{3}{c|}{Entity Extraction} & \multicolumn{3}{c}{Entity Linking} \\
\multicolumn{1}{c|}{Model} & \multicolumn{1}{c}{F} & \multicolumn{1}{c}{S} & \multicolumn{1}{c}{C} & \multicolumn{1}{|c}{F} & \multicolumn{1}{c}{C} & \multicolumn{1}{c}{Sci} \\

\toprule
BERT$_{\text{BASE}}$            & 60.92 & 93.67 & 93.13 & 27.65 & 92.83 & 86.76 \\
LayoutLM$^*_{\text{BASE}}$      & 78.54 & 95.11 & 96.26 & 45.86 & 95.21 & 99.05 \\
LayoutLMv2$^*_{\text{BASE}}$    & 81.89 & 96.09 & 96.05 & 42.91 & 95.59 & 98.19 \\
BROS$_{\text{BASE}}$            & \textbf{83.05} & \textbf{96.28} & \textbf{96.50} & \textbf{71.46} & \textbf{95.73} & \textbf{99.45} \\
\midrule
\midrule
BERT$_{\text{LARGE}}$           & 64.17 & 94.25 & 94.74 & 29.11 & 94.31 & 89.23 \\
LayoutLM$^*_{\text{LARGE}}$     & 79.27 & 95.36 & 96.12 & 42.83 & 95.41 & 99.33 \\
LayoutLMv2$^*_{\text{LARGE}}$   & 83.59 & 96.39 & 97.24 & 70.57 & 97.29 & \textbf{99.76} \\
BROS$_{\text{LARGE}}$           & \textbf{84.52} & \textbf{96.62} & \textbf{97.28} & \textbf{77.01} & \textbf{97.40} & 99.58 \\
\bottomrule

\end{tabular}
\end{adjustbox}
\caption{Performance comparisons on three EE and EL tasks \underline{\textit{with}} the order information of text blocks.}
\label{tbl:w_order_ee_and_el}
\vspace{-0.5em}
\end{table}

Table~\ref{tbl:w_order_ee_and_el} shows the F1 scores on three EE and EL tasks with the order of text blocks given in the dataset.
F, S, C, and Sci refer to FUNSD, SROIE$^*$, CORD, and SciTSR, respectively.
For EE tasks, all models utilize BIO tagger that captures spans of text blocks to represent key entities in documents.
For EL tasks, SPADE decoder is used to identify relationships between entities not placed sequentially in a series of text blocks. 
In all cases, BERT performs the worst because those tasks require understanding texts in 2D space, but BERT only encodes 1D sequential information.
LayoutLM$^*$ and LayoutLMv2$^*$ show better performance than BERT since they encode layout features as well as text features.
And by combining visual features, LayoutLMv2$^*$ performs better than LayoutLM$^*$ in most tasks.
BROS shows the best performance in all tasks except SciTSR.
It should be noted that the BROS$_{\text{BASE}}$ show better performance than that of LayoutLM$^*_{\text{LARGE}}$, even though it uses three times lower number of parameters (110M vs 343M).
These results indicate that BROS effectively encodes the text and layout features.

\begin{table}[t!]
\tabcolsep = 3pt
\centering
\begin{adjustbox}{width=23.5em}
\begin{tabular}{l|rrr|rrr}
 & \multicolumn{3}{c|}{Entity Extraction} & \multicolumn{3}{c}{Entity Linking} \\
\multicolumn{1}{c|}{Model} & \multicolumn{1}{c}{p-F} & \multicolumn{1}{c}{p-S} & \multicolumn{1}{c}{p-C} & \multicolumn{1}{|c}{p-F} & \multicolumn{1}{c}{p-C} & \multicolumn{1}{c}{p-Sci} \\

\toprule
BERT$_{\text{BASE}}$            & 18.85 & 39.73 & 59.71 &  9.59 & 27.88 &  1.75 \\
LayoutLM$^*_{\text{BASE}}$      & 33.89 & 66.05 & 80.86 & 22.98 & 61.51 & 97.32 \\
LayoutLMv2$^*_{\text{BASE}}$    & 40.77 & 73.56 & 80.37 & 23.25 & 50.55 & 95.86 \\
BROS$_{\text{BASE}}$            & \textbf{76.94} & \textbf{82.85} & \textbf{95.86} & \textbf{69.61} & \textbf{87.72} & \textbf{99.19} \\
\midrule
\midrule
BERT$_{\text{LARGE}}$           & 18.10 & 43.19 & 57.17 & 10.81 & 27.12 &  1.93 \\
LayoutLM$^*_{\text{LARGE}}$     & 33.11 & 56.84 & 82.88 & 20.72 & 61.98 & 97.64 \\
LayoutLMv2$^*_{\text{LARGE}}$   & 62.53 & 84.92 & 94.43 & 50.14 & 85.80 & \textbf{99.45} \\
BROS$_{\text{LARGE}}$           & \textbf{79.42} & \textbf{85.14} & \textbf{96.81} & \textbf{75.61} & \textbf{90.49} & 99.33 \\
\bottomrule

\end{tabular}
\end{adjustbox}
\caption{Performance comparisons on three EE and EL tasks \underline{\textit{without}} the order information of text blocks.}
\label{tbl:wo_order_ee_and_el}
\vspace{-0.5em}
\end{table}

\begin{table}[t!]
    \centering
    \begin{adjustbox}{width=23.50em}
    \begin{tabular}{l|cccc}
    \multicolumn{1}{c|}{Model}   & \multicolumn{1}{c}{p-}   & \multicolumn{1}{c}{xy-}  & \multicolumn{1}{c}{yx-}  & \multicolumn{1}{c}{original} \\
    \toprule
    LayoutLM$^*_{\text{BASE}}$          & 33.89 & 34.02 & 55.47 & 78.47 \\
    LayoutLMv2$^*_{\text{BASE}}$          & 40.77 & 52.08 & 62.37 & 78.16 \\
    BROS$_{\text{BASE}}$                & \textbf{76.94} & \textbf{77.16} & \textbf{77.42} & \textbf{81.61} \\
    \midrule
    \midrule
    LayoutLM$^*_{\text{LARGE}}$          & 33.11 & 33.54 & 41.45 & 48.30 \\
    LayoutLMv2$^*_{\text{LARGE}}$        & 62.53 & 69.14 & 75.45 & 83.00 \\
    BROS$_{\text{LARGE}}$                & \textbf{79.42} & \textbf{79.91} & \textbf{80.02} & \textbf{83.23} \\
    \bottomrule
    \end{tabular}
    \end{adjustbox}
    \caption{Comparison of FUNSD EE performances according to sorting methods. %
    }
    \label{tbl:tbl_sorting_funsd_ee}
    \vspace{-0.5em}
\end{table}

\subsubsection{Without the Order Information of Text Blocks}

As we mentioned in previous section, we introduce the permuted KIE benchmarks lost the orders of text blocks by shuffling the provided orders.
To solve EE and EL tasks without the order information, we employ the SPADE decoder for all tasks.
Table~\ref{tbl:wo_order_ee_and_el} shows the comparison results. %
p-F, p-S, p-C, and p-Sci refer to p-FUNSD, p-SROIE$^*$, p-CORD, and p-SciTSR, respectively.
BERT, which does not employ any spatial information of text blocks, shows the worst results on the orderless conditions.  
By being aware of the spatiality, layout-aware language models show better performances than BERT, and BROS achieves the best except p-SciTSR. 
More interestingly, BROS shows minor performance drops compared to Table~\ref{tbl:w_order_ee_and_el}, while LayoutLM$^*$ and LayoutLMv2$^*$ suffer from huge performance degradations by losing the order information of text blocks.

To systematically investigate how the order information affects the performance of the models, we construct variants of FUNSD by re-ordering text blocks with two sorting methods based on the top-left points.
The text blocks of xy-FUNSD are sorted according to the x-axis with ascending order of y-axis and those of yx-FUNSD are sorted according to y-axis with ascending order of x-axis.
Table~\ref{tbl:tbl_sorting_funsd_ee} shows performance on p-FUNSD, xy-FUNSD, yx-FUNSD, and the original FUNSD with the SPADE decoder.
In our experiment, LayoutLM$^*_{\text{LARGE}}$ achieves unstable performance (48.30$\pm$12.51), when combined with SPADE decoder.
Interestingly, the performances of LayoutLM$^*$ and LayoutLMv2$^*$ are degraded in the order of FUNSD, yx-FUNSD, xy-FUNSD, and p-FUNSD as like the order of the reasonable serialization for text on 2D space.
On the other hand, the performance of BROS is relatively consistent.
These results show the robustness of BROS on multiple types of serializers.

\begin{figure}
    \centering
    \begin{subfigure}{.495\linewidth}
        \centering
        \includegraphics[width=\linewidth]{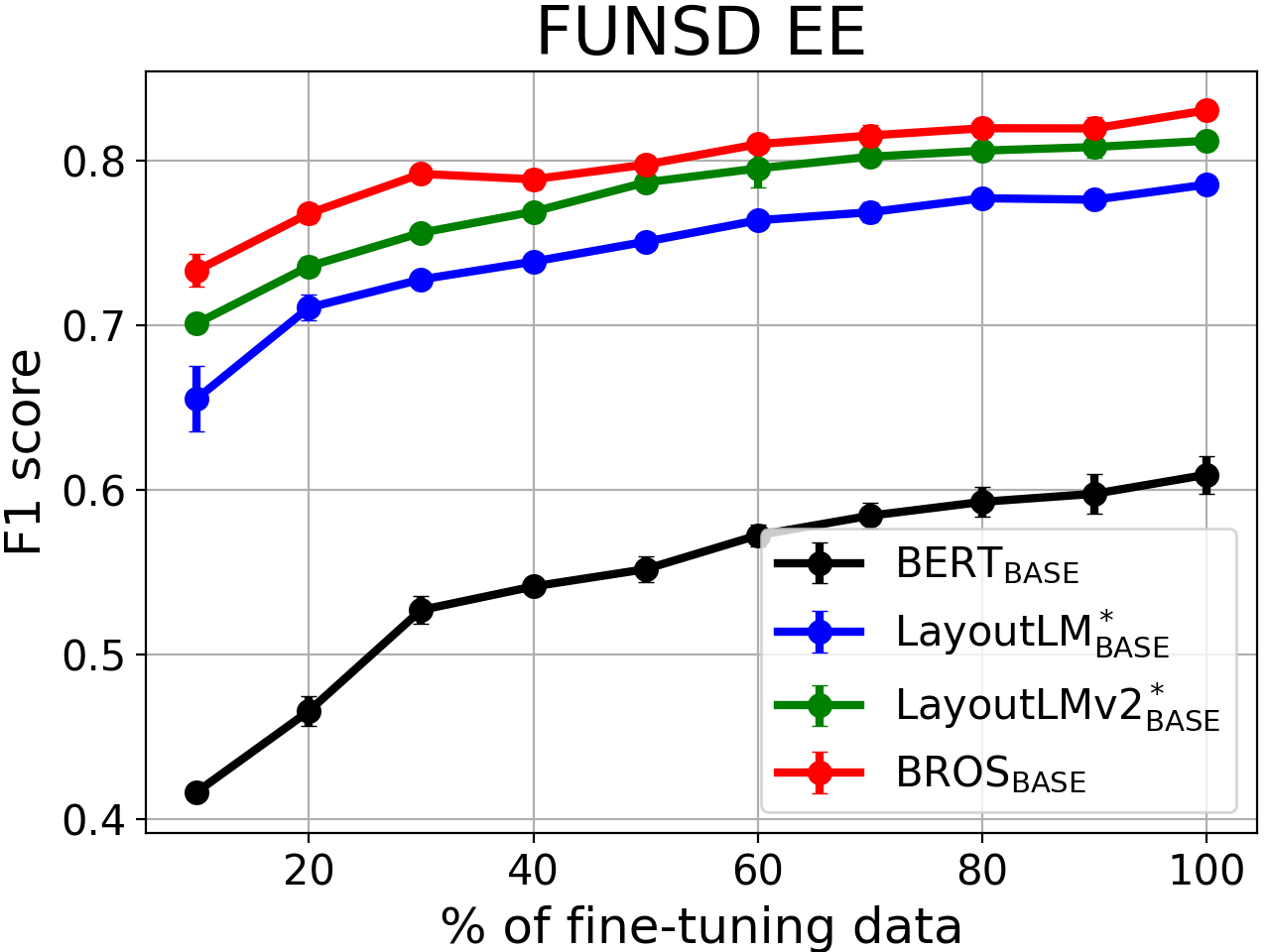}
    \end{subfigure}
    \begin{subfigure}{.495\linewidth}
        \centering
        \includegraphics[width=\linewidth]{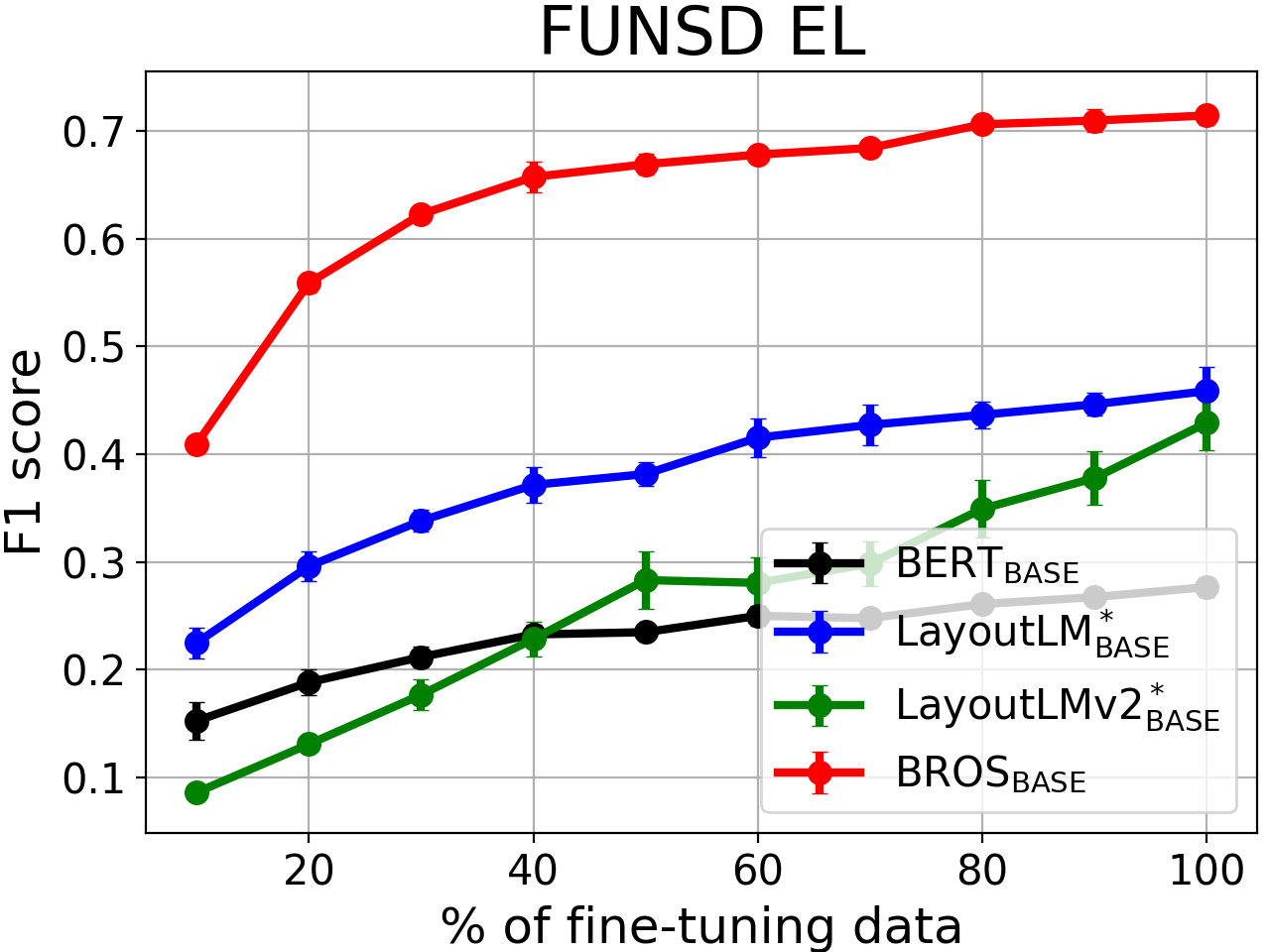}
    \end{subfigure}
    \caption{Performance comparisons according to the amount of fine-tuning data. Each point represents the result of fine-tuning using from 10\% to 100\% of training data.}
    \label{fig:fine_tuning_ratio}
    \vspace{-0.5em}
\end{figure}

\subsubsection{Learning from Few Training Examples}

\begin{table}[t]
\tabcolsep = 3pt
\centering
\begin{adjustbox}{width=23.5em}
\begin{tabular}{c|r|cccc}
Dataset & \multicolumn{1}{c|}{\# Data} & \multicolumn{1}{c}{\small{BERT}} & \multicolumn{1}{c}{\small{LayoutLM$^*$}} & \multicolumn{1}{c}{\small{LayoutLMv2$^*$}} & \multicolumn{1}{c}{\small{BROS}} \\

\toprule
FUNSD       &  5 & 31.51 & 48.23 & 64.26 & \textbf{68.35} \\
EE          & 10 & 40.46 & 62.50 & 69.92 & \textbf{72.60} \\
\midrule
FUNSD       &  5 & 14.65 & 21.38 &  7.32 & \textbf{31.11} \\
EL          & 10 & 14.88 & 21.48 & 13.99 & \textbf{39.17} \\
\bottomrule

\end{tabular}
\end{adjustbox}
\caption{Results of training with 5 and 10 examples.}
\label{tbl:few_shot}
\vspace{-0.6em}
\end{table}

\begin{table}[t]
\tabcolsep = 3pt
\centering
\begin{adjustbox}{width=23.5em}
\begin{tabular}{l|rrr|rrr}
& \multicolumn{3}{c|}{Entity Extraction} & \multicolumn{3}{c}{Entity Linking} \\
\multicolumn{1}{c|}{Model} & \multicolumn{1}{c}{F} & \multicolumn{1}{c}{S} & \multicolumn{1}{c|}{C} & \multicolumn{1}{c}{F} & \multicolumn{1}{c}{C} & \multicolumn{1}{c}{Sci} \\

\toprule
LayoutLM$^\dagger_{\textbf{BASE}}$          & 76.89 & 94.99 & 94.37 & 44.00 & 93.60 & 99.06 \\
\midrule
\;$\rightarrow$ \textit{pos enc. only}      & 78.84 & 95.45 & 96.36 & 59.92 & 94.83 & 99.22 \\
\;$\rightarrow$ \textit{objectives only}    & 78.44 & 94.81 & 95.95 & 47.22 & 94.11 & 99.20 \\
\midrule
\;$\rightarrow$ \textit{both} (= $\textbf{BROS}_{\textbf{BASE}}$) & \textbf{80.58} & \textbf{95.72} & \textbf{96.64} & \textbf{65.24} & \textbf{96.03} & \textbf{99.28} \\
\bottomrule

\end{tabular}
\end{adjustbox}
\caption{Performance improvements on EE and EL tasks through adding components of BROS. At the last line, all components are changed from LayoutLM and the model becomes BROS.}
\label{tbl:ablation_grad_ee_and_el}
\vspace{-0.5em}
\end{table}

\begin{table}[t]
\tabcolsep = 2.5pt
\centering
\begin{adjustbox}{width=23.5em}
\begin{tabular}{l|rrr|rrr}
& \multicolumn{3}{c|}{Entity Extraction} & \multicolumn{3}{c}{Entity Linking} \\
\multicolumn{1}{c|}{SE} & \multicolumn{1}{c}{F} & \multicolumn{1}{c}{S} & \multicolumn{1}{c|}{C} & \multicolumn{1}{c}{F} & \multicolumn{1}{c}{C} & \multicolumn{1}{c}{Sci} \\

\toprule
Absolute & 78.44 & 94.81 & 95.95 & 47.22 & 94.11 & 99.20 \\
LayoutLMv2's & 78.93 & 94.71 & 95.82 & 53.57 & 95.27 & \textbf{99.28} \\
\midrule
Ours & \textbf{80.58} & \textbf{95.72} & \textbf{96.64} & \textbf{65.24} & \textbf{96.03} & \textbf{99.28} \\
\bottomrule

\end{tabular}
\end{adjustbox}
\caption{
Spatial encoding methods from BROS' setting.
}
\label{tbl:ablation_spatial_encoding}
\vspace{-0.5em}
\end{table}

One of the advantages of pre-trained models is that it shows effective transfer learning performance even with a few training examples~\cite{BERT}.
Since collecting fine-tuning data requires a lot of resource, achieving high performance with a small number of training examples is important.

Figure~\ref{fig:fine_tuning_ratio} shows the results of the FUNSD KIE tasks by varying the amount of training examples from 10\% to 100\% during fine-tuning.
In all models, performances tend to increase as the ratio of training data increased.
In both tasks, BROS shows the best performances regardless of the number of training samples.

To further test extreme cases, we conduct experiments using only 5 and 10 training examples.
Table~\ref{tbl:few_shot} shows the results of the FUNSD KIE tasks.
We fine-tune models for 100 epochs with a batch size of 4.
In all cases, BROS shows the best performances.
The results prove the generalization ability of BROS even with very few training examples.

\subsection{Ablation Study}

We conduct ablation studies to investigate which component contributes the performance improvement.
For the ablation studies, we utilize LayoutLM$^\dagger$ that is our own implementation of LayoutLM for fair comparisons under the same experimental settings.
All models in these studies are pre-trained for 1 epoch. 

Table~\ref{tbl:ablation_grad_ee_and_el} provides performance changes from adding our proposed components. 
When applying our proposed positional encoding to LayoutLM, the performances consistently increase with huge margins of 3.62pp on average over all tasks.
Independently, our extension on pre-training objectives solely provides 1.14pp of performance improvement on average.
By utilizing both, BROS$_{\text{BASE}}$ provides the best performances with margins of 5.10pp on average.
This ablation study proves that each component of BROS solely contributes to performance improvements as well as their combination provides better results.

Table~\ref{tbl:ablation_spatial_encoding} compares three positional encoding methods: absolute position in LayoutLM, relative position in LayoutLMv2, and ours.
Relative position methods perform better than absolute one and the performance gap becomes larger in EL tasks.
And among them, our method shows the best results.

\section{Conclusion}

We propose a pre-trained language model, BROS, which focuses on modeling text and layout features for effective key information extraction from documents.
By encoding texts in 2D space with their relative positions and pre-training the model with the area-masking strategy, BROS shows superior performance without relying on any additional visual features.
In addition, under the two real-world settings--imprecise text serialization and small amount of training examples--BROS shows robust performance while other models show significant performance degradation.

\section{Acknowledgments}

We thank many colleagues at NAVER CLOVA for their help, in particular Yoonsik Kim, Moonbin Yim, Han-cheol Cho, Bado Lee, Seunghyun Park, and Youngmin Baek for useful discussions.

\bibliography{references}

\begin{thebibliography}{25}
\providecommand{\natexlab}[1]{#1}

\bibitem[{Appalaraju et~al.(2021)Appalaraju, Jasani, Kota, Xie, and
  Manmatha}]{DocFormer}
Appalaraju, S.; Jasani, B.; Kota, B.~U.; Xie, Y.; and Manmatha, R. 2021.
\newblock DocFormer: End-to-End Transformer for Document Understanding.
\newblock \emph{arXiv preprint arXiv:2106.11539}.

\bibitem[{Chi et~al.(2019)Chi, Huang, Xu, Yu, Yin, and
  Mao}]{chi2019complicated}
Chi, Z.; Huang, H.; Xu, H.-D.; Yu, H.; Yin, W.; and Mao, X.-L. 2019.
\newblock Complicated table structure recognition.
\newblock \emph{arXiv preprint arXiv:1908.04729}.

\bibitem[{Clausner, Pletschacher, and
  Antonacopoulos(2013)}]{clausner2013significance}
Clausner, C.; Pletschacher, S.; and Antonacopoulos, A. 2013.
\newblock The significance of reading order in document recognition and its
  evaluation.
\newblock In \emph{2013 12th International Conference on Document Analysis and
  Recognition (ICDAR)}, 688--692. IEEE.

\bibitem[{Dai et~al.(2019)Dai, Yang, Yang, Carbonell, Le, and
  Salakhutdinov}]{TransformerXL}
Dai, Z.; Yang, Z.; Yang, Y.; Carbonell, J.~G.; Le, Q.; and Salakhutdinov, R.
  2019.
\newblock Transformer-XL: Attentive Language Models beyond a Fixed-Length
  Context.
\newblock In \emph{Proceedings of the 57th Annual Meeting of the Association
  for Computational Linguistics (ACL)}.

\bibitem[{Denk and Reisswig(2019)}]{BERTgrid}
Denk, T.~I.; and Reisswig, C. 2019.
\newblock {BERT}grid: Contextualized Embedding for 2D Document Representation
  and Understanding.
\newblock In \emph{Workshop on Document Intelligence at NeurIPS 2019}.

\bibitem[{Devlin et~al.(2019)Devlin, Chang, Lee, and Toutanova}]{BERT}
Devlin, J.; Chang, M.-W.; Lee, K.; and Toutanova, K. 2019.
\newblock {BERT}: Pre-training of Deep Bidirectional Transformers for Language
  Understanding.
\newblock In \emph{Proceedings of the 2019 Conference of the North American
  Chapter of the Association for Computational Linguistics: Human Language
  Technologies (NAACL-HLT), Volume 1 (Long and Short Papers)}, 4171--4186.

\bibitem[{Harley, Ufkes, and Derpanis(2015)}]{harley2015evaluation}
Harley, A.~W.; Ufkes, A.; and Derpanis, K.~G. 2015.
\newblock Evaluation of deep convolutional nets for document image
  classification and retrieval.
\newblock In \emph{Proceedings of the 13th International Conference on Document
  Analysis and Recognition (ICDAR)}, 991--995.

\bibitem[{Huang et~al.(2019)Huang, Chen, He, Bai, Karatzas, Lu, and
  Jawahar}]{huang2019icdar2019}
Huang, Z.; Chen, K.; He, J.; Bai, X.; Karatzas, D.; Lu, S.; and Jawahar, C.
  2019.
\newblock {ICDAR}2019 competition on scanned receipt ocr and information
  extraction.
\newblock In \emph{Proceedings of the 15th International Conference on Document
  Analysis and Recognition (ICDAR)}, 1516--1520. IEEE.

\bibitem[{Hwang et~al.(2019)Hwang, Kim, Seo, Yim, Park, Park, Lee, Lee, and
  Lee}]{POST}
Hwang, W.; Kim, S.; Seo, M.; Yim, J.; Park, S.; Park, S.; Lee, J.; Lee, B.; and
  Lee, H. 2019.
\newblock {Post-OCR} parsing: building simple and robust parser via BIO
  tagging.
\newblock In \emph{Workshop on Document Intelligence at NeurIPS 2019}.

\bibitem[{Hwang et~al.(2021)Hwang, Yim, Park, Yang, and Seo}]{SPADE}
Hwang, W.; Yim, J.; Park, S.; Yang, S.; and Seo, M. 2021.
\newblock Spatial Dependency Parsing for Semi-Structured Document Information
  Extraction.
\newblock In \emph{Findings of the Association for Computational Linguistics:
  ACL-IJCNLP 2021}, 330--343.

\bibitem[{Jaume, Ekenel, and Thiran(2019)}]{jaume2019funsd}
Jaume, G.; Ekenel, H.~K.; and Thiran, J.-P. 2019.
\newblock {FUNSD}: A dataset for form understanding in noisy scanned documents.
\newblock In \emph{2019 International Conference on Document Analysis and
  Recognition Workshops (ICDARW)}, volume~2, 1--6. IEEE.

\bibitem[{Joshi et~al.(2020)Joshi, Chen, Liu, Weld, Zettlemoyer, and
  Levy}]{SpanBERT}
Joshi, M.; Chen, D.; Liu, Y.; Weld, D.~S.; Zettlemoyer, L.; and Levy, O. 2020.
\newblock Span{BERT}: Improving pre-training by representing and predicting
  spans.
\newblock \emph{Transactions of the Association for Computational Linguistics
  (TACL)}, 8: 64--77.

\bibitem[{Lewis et~al.(2006)Lewis, Agam, Argamon, Frieder, Grossman, and
  Heard}]{Lewis2006building}
Lewis, D.; Agam, G.; Argamon, S.; Frieder, O.; Grossman, D.; and Heard, J.
  2006.
\newblock Building a test collection for complex document information
  processing.
\newblock In \emph{Proceedings of the 29th annual international ACM SIGIR
  conference on Research and development in information retrieval (SIGIR)},
  665--666.

\bibitem[{Li et~al.(2021{\natexlab{a}})Li, Bi, Yan, Wang, Huang, Huang, and
  Si}]{StructuralLM}
Li, C.; Bi, B.; Yan, M.; Wang, W.; Huang, S.; Huang, F.; and Si, L.
  2021{\natexlab{a}}.
\newblock StructuralLM: Structural Pre-training for Form Understanding.
\newblock In \emph{Proceedings of the 59th Annual Meeting of the Association
  for Computational Linguistics and the 11th International Joint Conference on
  Natural Language Processing (ACL-IJCNLP)}, 6309--6318.

\bibitem[{Li et~al.(2020)Li, Gao, Bu, Wang, Yu, and Zheng}]{ECCV-ORDER}
Li, L.; Gao, F.; Bu, J.; Wang, Y.; Yu, Z.; and Zheng, Q. 2020.
\newblock An End-to-End {OCR} Text Re-organization Sequence Learning for
  Rich-text Detail Image Comprehension.
\newblock In \emph{Proceedings of the 16th European Conference on Computer
  Vision (ECCV)}.

\bibitem[{Li et~al.(2021{\natexlab{b}})Li, Gu, Kuen, Morariu, Zhao, Jain,
  Manjunatha, and Liu}]{SelfDoc}
Li, P.; Gu, J.; Kuen, J.; Morariu, V.~I.; Zhao, H.; Jain, R.; Manjunatha, V.;
  and Liu, H. 2021{\natexlab{b}}.
\newblock SelfDoc: Self-Supervised Document Representation Learning.
\newblock In \emph{Proceedings of the IEEE/CVF Conference on Computer Vision
  and Pattern Recognition}, 5652--5660.

\bibitem[{Li et~al.(2021{\natexlab{c}})Li, Qian, Yu, Qin, Zhang, Liu, Yao, Han,
  Liu, and Ding}]{StrucTexT}
Li, Y.; Qian, Y.; Yu, Y.; Qin, X.; Zhang, C.; Liu, Y.; Yao, K.; Han, J.; Liu,
  J.; and Ding, E. 2021{\natexlab{c}}.
\newblock StrucTexT: Structured Text Understanding with Multi-Modal
  Transformers.
\newblock \emph{arXiv preprint arXiv:2108.02923}.

\bibitem[{Liu et~al.(2019)Liu, Gao, Zhang, and Zhao}]{liu2019graph}
Liu, X.; Gao, F.; Zhang, Q.; and Zhao, H. 2019.
\newblock Graph Convolution for Multimodal Information Extraction from Visually
  Rich Documents.
\newblock In \emph{Proceedings of the 2019 Conference of the North American
  Chapter of the Association for Computational Linguistics: Human Language
  Technologies (NAACL-HLT), Volume 2 (Industry Papers)}, 32--39.

\bibitem[{Loshchilov and Hutter(2019)}]{AdamW}
Loshchilov, I.; and Hutter, F. 2019.
\newblock Decoupled Weight Decay Regularization.
\newblock In \emph{Proceedings of the 7th International Conference on Learning
  Representations (ICLR)}.

\bibitem[{Park et~al.(2019)Park, Shin, Lee, Lee, Surh, Seo, and
  Lee}]{park2019cord}
Park, S.; Shin, S.; Lee, B.; Lee, J.; Surh, J.; Seo, M.; and Lee, H. 2019.
\newblock {CORD}: A Consolidated Receipt Dataset for Post-OCR Parsing.
\newblock In \emph{Workshop on Document Intelligence at NeurIPS 2019}.

\bibitem[{Powalski et~al.(2021)Powalski, Borchmann, Jurkiewicz, Dwojak,
  Pietruszka, and Pa{\l}ka}]{TILT}
Powalski, R.; Borchmann, {\L}.; Jurkiewicz, D.; Dwojak, T.; Pietruszka, M.; and
  Pa{\l}ka, G. 2021.
\newblock Going Full-TILT Boogie on Document Understanding with
  Text-Image-Layout Transformer.
\newblock \emph{arXiv preprint arXiv:2102.09550}.

\bibitem[{Vaswani et~al.(2017)Vaswani, Shazeer, Parmar, Uszkoreit, Jones,
  Gomez, Kaiser, and Polosukhin}]{Transformer}
Vaswani, A.; Shazeer, N.; Parmar, N.; Uszkoreit, J.; Jones, L.; Gomez, A.~N.;
  Kaiser, {\L}.; and Polosukhin, I. 2017.
\newblock Attention is all you need.
\newblock In \emph{Advances in Neural Information Processing Systems 30
  (NeurIPS)}, 5998--6008.

\bibitem[{Wang et~al.(2021)Wang, Xu, Cui, Shang, and Wei}]{LayoutReader}
Wang, Z.; Xu, Y.; Cui, L.; Shang, J.; and Wei, F. 2021.
\newblock LayoutReader: Pre-training of Text and Layout for Reading Order
  Detection.
\newblock arXiv:2108.11591.

\bibitem[{Xu et~al.(2020)Xu, Li, Cui, Huang, Wei, and Zhou}]{LayoutLM}
Xu, Y.; Li, M.; Cui, L.; Huang, S.; Wei, F.; and Zhou, M. 2020.
\newblock {LayoutLM}: Pre-training of text and layout for document image
  understanding.
\newblock In \emph{Proceedings of the 26th ACM SIGKDD International Conference
  on Knowledge Discovery \& Data Mining (KDD)}, 1192--1200.

\bibitem[{Xu et~al.(2021)Xu, Xu, Lv, Cui, Wei, Wang, Lu, Florencio, Zhang, Che
  et~al.}]{LayoutLMv2}
Xu, Y.; Xu, Y.; Lv, T.; Cui, L.; Wei, F.; Wang, G.; Lu, Y.; Florencio, D.;
  Zhang, C.; Che, W.; et~al. 2021.
\newblock LayoutLMv2: Multi-modal Pre-training for Visually-Rich Document
  Understanding.
\newblock In \emph{Proceedings of the 59th Annual Meeting of the Association
  for Computational Linguistics and the 11th International Joint Conference on
  Natural Language Processing (ACL-IJCNLP)}, 2579--2591.

\end{thebibliography}

\clearpage

\onecolumn

\appendix

\section{Two Categories of Document KIE tasks and Parsers for Them}

Document KIE tasks can be categorized into two downstream tasks: (1) an entity extraction (EE) task and (2) an entity linking (EL) task.
The EE task identifies sequences of text blocks that represent desired target texts (e.g. extracts menu name in a receipt) and the EL task connects key entities through their hierarchical or semantic relations (e.g. connects menu name and its price).

To address the EE and EL tasks, we introduce two parsers.
One is a sequence classifier, BIO tagger, that can operate based on prior information about an order of text blocks.
It has been utilized as a conventional method for extracting entities from the text sequences.
The other is a graph-based parser, SPADE~\cite{SPADE} decoder, that does not require any order information of text blocks.
Since it directly identifying relations between tokens, it can extract entities whether the order of text blocks is provided or not.
It can also perform the task which requires inferring the relationship between text blocks.
The following sections briefly introduce the two types of parsers.

\subsection{BIO Tagger}

BIO tagger, a representative parser that depends on the proper order of text blocks, extracts key information by identifying the beginning (B) and inside (I) points of the ordered text blocks.
The proper order of text blocks indicates an order in which all of the key information can be represented in its sub-sequences.
The sequence classifier requires this condition because it never succeeds to find the key information with a wrong sequence.
For example, if three text blocks, ``optical'', ``character'', and ``recognition'', are ordered as ``recognition'', ``optical'', and ``character'', the sequence classifier cannot find ``optical character recognition''.

\begin{figure*}[h]
    \centering
    \begin{subfigure}[t]{.325\textwidth}
        \centering
        \includegraphics[width=0.95\linewidth]{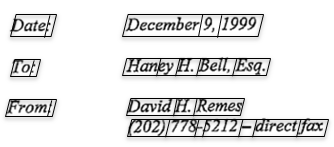}
        \caption{Recognized text blocks.}
    \end{subfigure}
    \begin{subfigure}[t]{.325\textwidth}
        \centering
        \includegraphics[width=0.95\linewidth]{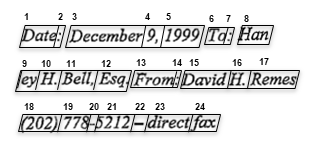}
        \caption{Serialized text blocks.}
    \end{subfigure}
    \begin{subfigure}[t]{.325\textwidth}
        \centering
        \includegraphics[width=0.95\linewidth]{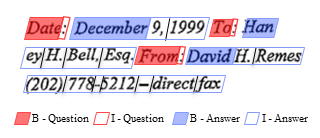}
        \caption{BIO-tagged text sequence.}
    \end{subfigure}
    \caption{Visual descriptions of how BIO tagger extracts entities in a document. All recognized tokens are serialized and classified. By combining sub-sequences identified by the BIO taggings, key information can be parsed from the recognized tokens.}
    \label{fig:bio_tagger}
\end{figure*}

Figure~\ref{fig:bio_tagger} shows how the BIO tagger performs the EE task for a given document.
First, text blocks are recognized by an OCR engine (Figure~\ref{fig:bio_tagger}, a).
The recognized text blocks are then serialized by a serializer (Figure~\ref{fig:bio_tagger}, b).
Finally, for each token, the BIO classes are classified and key information is extracted by combining the classified labels (Figure~\ref{fig:bio_tagger}, c).

The BIO tagger cannot solve the EL task since links between text blocks cannot be represented as a sequence unit.
In addition, a single text block can hold the same relationships with other multiple text blocks but the sequence-based approach cannot explain the one-to-many relations as well.

\subsection{SPADE Decoder}

In many practical cases, the proper order of text blocks cannot be available.
Most OCR APIs provide the order of text blocks based on rule-based approaches but they cannot guarantee the proper order of text blocks~\cite{clausner2013significance,ECCV-ORDER,LayoutReader}. 

Here, we utilize SPADE~\cite{SPADE} decoder to extract key information without any information about the order. 
The key idea of SPADE decoder is to extract a directional sub-graph from a fully-connected graph which nodes are text blocks.
Due to no limitation on the connections between text blocks, it does not require order information.
In our paper, we slightly modified the SPADE decoder for its application to the pre-trained models.
The details are described as the followings.

\begin{figure*}[h]
\captionsetup[subfigure]{aboveskip=1pt,belowskip=0pt}
    \centering
    \begin{subfigure}{.325\textwidth}
        \centering
        \includegraphics[width=0.95\linewidth]{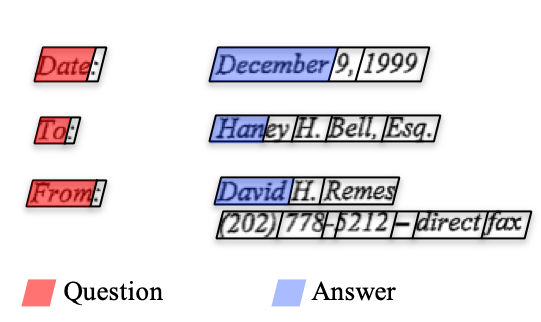}
        \caption{Initial token classification}
    \end{subfigure}
    \begin{subfigure}{.325\textwidth}
        \centering
        \includegraphics[width=0.95\linewidth]{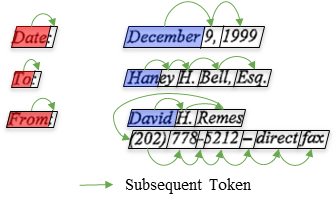}
        \caption{Subsequent token classification}
    \end{subfigure}
    \begin{subfigure}{.325\textwidth}
        \centering
        \includegraphics[width=0.95\linewidth]{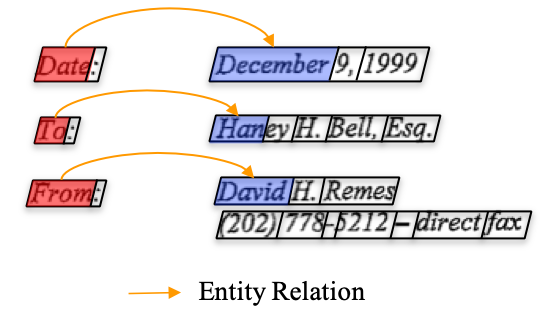}
        \caption{Entity linking (EL) task}
    \end{subfigure}
    \caption{Visual descriptions of SPADE decoder downstream tasks. For EE tasks, SPADE decoder combines two sub-tasks such as (a) and (b). SPADE decoder identifies initial tokens and then connects next tokens without any order information of text blocks. For the EL task, SPADE decoder links the first tokens of the entities.}
    \label{fig:task_description}
\end{figure*}

For EE tasks, the SPADE decoder divides the problem into two sub-tasks: initial token classification (Figure~\ref{fig:task_description}, a) and subsequent token classification (Figure~\ref{fig:task_description}, b).
Let $\tilde{\vt}_i \in \mathbb{R}^{H}$ denote the $i^{\text{th}}$ token representation from the last Transformer layer of the pre-trained model.
The initial token classification conducts a token-level tagging to determine whether a token is an initial token of target information as follows,
\begin{equation}
    p_{\text{itc}}(\tilde{\vt}_i) = \text{softmax}(\mW^{\text{itc}}\tilde{\vt}_i), %
\end{equation}
where $\mW^{\text{itc}} \in \mathbb{R}^{(C+1) \times H }$ is a linear transition matrix and $C$ indicates the number of target classes.
Here, the extra +1 dimension is considered to indicate non-initial tokens.

The subsequent token classification is conducted by utilizing pair-wise token representations as follows,
\begin{align}
\begin{split}
    &p_{\text{stc}}(\tilde{\vt}_i) = \text{softmax} ( (\mW^{\text{stc-s}}\tilde{\vt}_i)^\top \mathbf{T}^{\text{stc}})^\top, \\
    &\text{where } \mathbf{T}^{\text{stc}} = [ \vt^{\text{stc}}; \mW^{\text{stc-t}}\tilde{\vt}_1; \dots; \mW^{\text{stc-t}}\tilde{\vt}_N ]. \nonumber
\end{split}
\end{align}
Here, $\mW^{\text{stc-s}}, \mW^{\text{stc-t}} \in \mathbb{R}^{H^{\text{stc}} \times H}$ are linear transition matrices, $H^{\text{stc}}$ is a hidden feature dimension for the subsequent token classification decoder and $N$ is the maximum number of tokens.
The semicolon (;) indicates concatenation.
$\vt^{\text{stc}} \in \mathbb{R}^{H^{\text{stc}}}$ is a model parameter to classify tokens which do not have a subsequent token or are not related to any class. It has a similar role with an end-of-sequence token, \texttt{[EOS]}, in NLP.
By solving these two sub-tasks, the SPADE decoder can identify a sequence of text blocks by finding initial tokens and then connecting subsequent tokens.

For EL tasks, the SPADE decoder conducts a binary classification for all possible pairs of tokens (Figure~\ref{fig:task_description}, c) as follows,
\begin{equation}
    p_{\text{rel}}(\tilde{\vt}_i, \tilde{\vt}_j) = \text{sigmoid}( (\mW^{\text{rel-s}}\tilde{\vt}_i)^\top (\mW^{\text{rel-t}}\tilde{\vt}_j) ),
\end{equation}
where $\mW^{\text{rel-s}}, \mW^{\text{rel-t}} \in \mathbb{R}^{H^{\text{rel}} \times H }$ are linear transition matrices and $H^{\text{rel}}$ is a hidden feature dimension.
Compared to the subsequent token classification, a single token can hold multiple relations with other tokens to represent hierarchical structures of document layouts.

In our experiments, $H^{\text{stc}}$ and $H^{\text{rel}}$ are set to 128 for FUNSD, 64 for SROIE$^*$, and 256 for CORD and SciTSR.

\begin{figure*}[!htb]
    \centering
    \begin{subfigure}{.45\textwidth}
        \centering
        \includegraphics[width=1.1\linewidth]{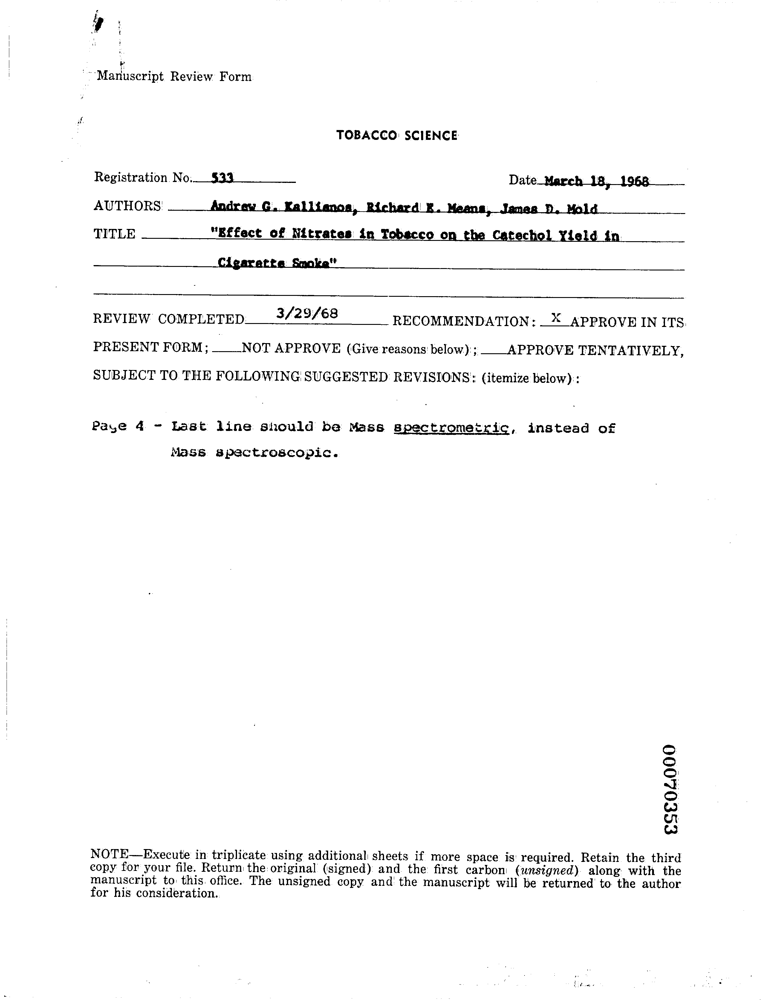}
        \caption{FUNSD}
    \end{subfigure}
    \begin{subfigure}{.45\textwidth}
        \centering
        \includegraphics[width=0.6\linewidth]{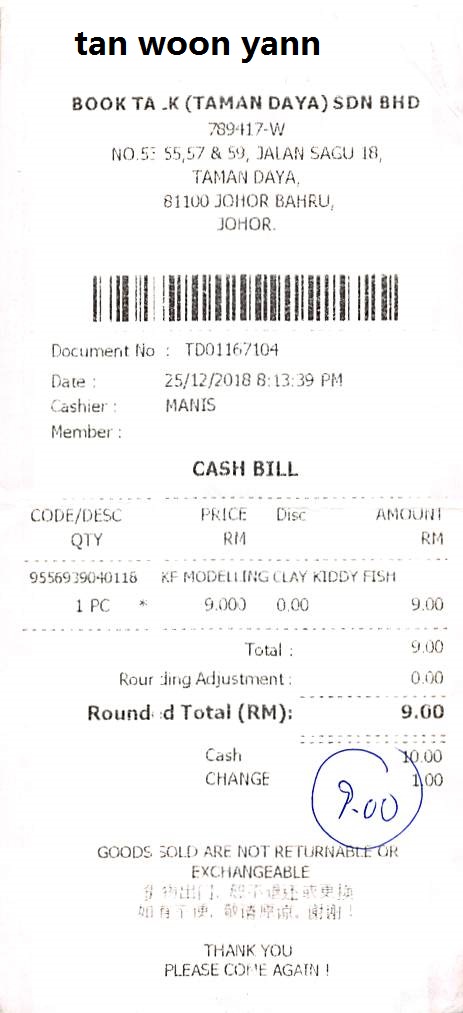}
        \caption{SROIE}
    \end{subfigure}
    \begin{subfigure}{.45\textwidth}
        \centering
        \includegraphics[width=0.6\linewidth]{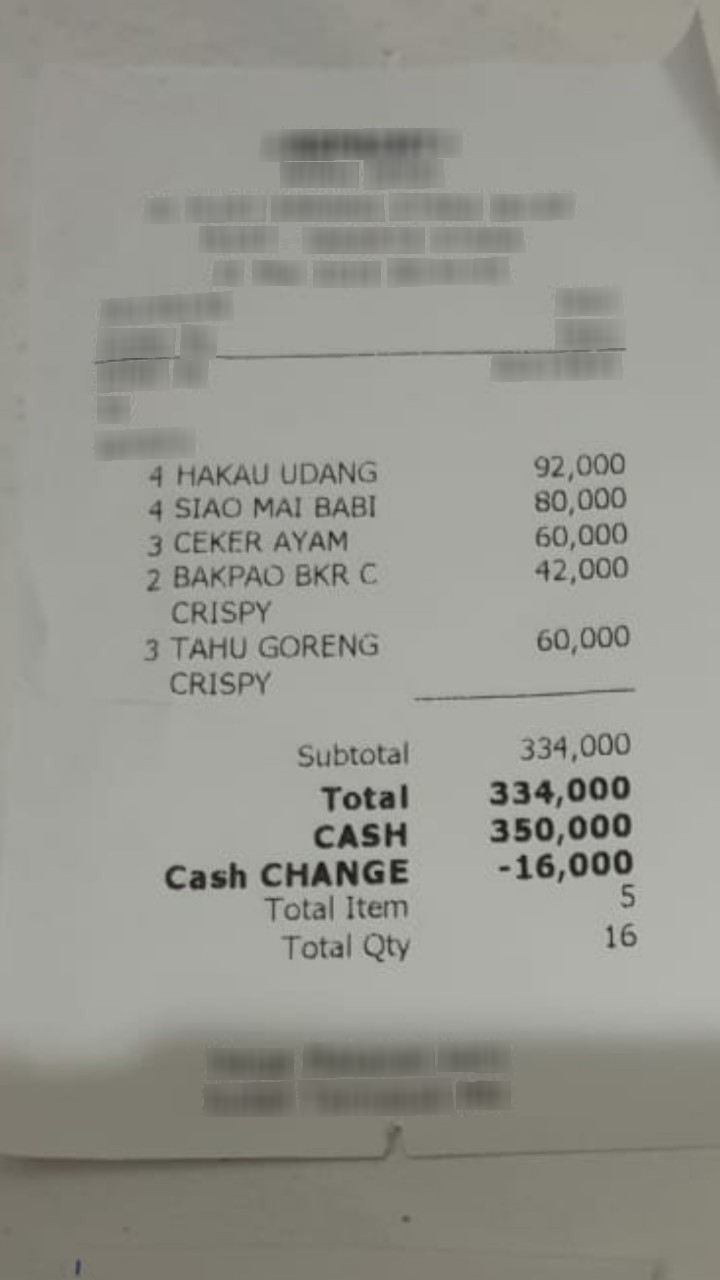}
        \caption{CORD}
    \end{subfigure}
    \begin{subfigure}{.45\textwidth}
        \centering
        \includegraphics[width=1.1\linewidth]{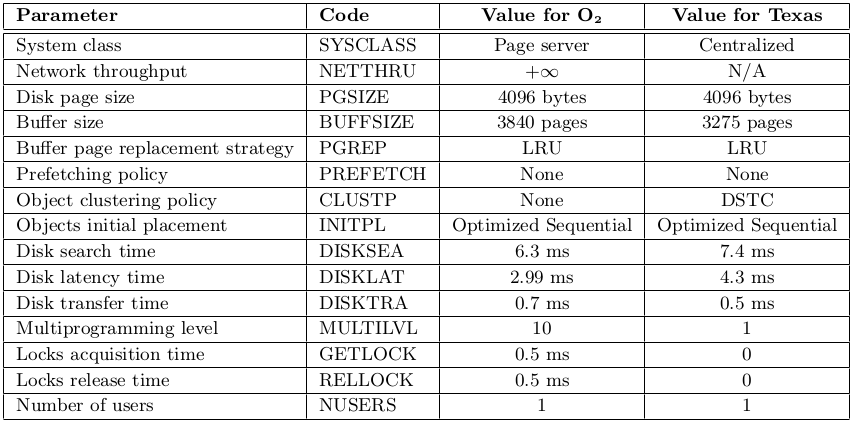}
        \caption{SciTSR}
    \end{subfigure}
    \caption{The sample images of four KIE benchmark datasets.}
    \label{fig:kie_samples}
\end{figure*}

\section{KIE Benchmark Datasets}

Here, we describe three EE tasks and three EL tasks from four KIE benchmark datasets.
\begin{itemize}[leftmargin=1.4em]
\setlength\itemsep{0.1em}
    \item Form Understanding in Noisy Scanned Documents (FUNSD)~\citep{jaume2019funsd} is a set of documents with various forms.
    The dataset consists of 149 training and 50 testing examples.
    FUNSD has both EE and EL tasks.
    In the EE task, there are three semantic entities: Header, Question, and Answer.
    In the EL task, the semantic hierarchies are represented as relations between text blocks like header-question and question-answer pairs. 
    \item SROIE$^*$ is a variant of Task 3 of ``Scanned Receipts OCR and Information Extraction'' (SROIE)\footnote{https://rrc.cvc.uab.es/?ch=13} that consists of a set of store receipts.
    In the original SROIE task, semantic contents (Company, Date, Address, and Total price) are generated without explicit connection to the text blocks.
    To convert SROIE into a EE task, we developed SROIE$^*$ by matching ground truth contents with text blocks.
    We also split the original training set into 526 training and 100 testing examples because the ground truths are not given in the original test set.
    SROIE$^*$ will be publicly available.
    \item Consolidated Receipt Dataset (CORD) \citep{park2019cord} is a set of store receipts with 800 training, 100 validation, and 100 testing examples.
    CORD consists of both EE and EL tasks.
    In the EE task, there are 30 semantic entities including menu name, menu price, and so on.
    In the EL task, the semantic entities are linked according to their layout structure.
    For example, menu name entities are linked to menu id, menu count, and menu price. 
    \item Complicated Table Structure Recognition (SciTSR) \citep{chi2019complicated} is an EL task that connects cells in a table to recognize the table structure.
    There are two types of relations: vertical and horizontal connections between cells.
    The dataset consists of 12,000 training images and 3,000 test images.
\end{itemize}

Figure~\ref{fig:kie_samples} shows the sample images of these benchmark datasets.

\clearpage

\begin{table}[t]

\centering

\begin{tabular}{rrl|rrr}
\multicolumn{1}{c}{\# Pre-training data} & \multicolumn{1}{c}{\# Epochs} & \multicolumn{1}{c}{Model} & \multicolumn{1}{c}{Precision} & \multicolumn{1}{c}{Recall} & \multicolumn{1}{c}{F1}   \\ \toprule
\multirow{2}{*}{500K} & \multirow{2}{*}{1}  & $\text{LayoutLM}_{\text{BASE}}$ \citep{LayoutLM} & 0.5779 & 0.6955 & 0.6313 \\
                      &                     & $\text{LayoutLM}^{\dagger}_{\text{BASE}}$     & 0.5823 & 0.6935 & 0.6330 \\ \midrule
\multirow{2}{*}{1M} & \multirow{2}{*}{1}    & $\text{LayoutLM}_{\text{BASE}}$ \citep{LayoutLM} & 0.6156 & 0.7005 & 0.6552 \\
                    &    & $\text{LayoutLM}^{\dagger}_{\text{BASE}}$     & 0.6142 & 0.7151 & 0.6608 \\ \midrule
\multirow{2}{*}{2M} & \multirow{2}{*}{1}     & $\text{LayoutLM}_{\text{BASE}}$ \citep{LayoutLM} & 0.6599 & 0.7355 & 0.6957 \\
                    &    & $\text{LayoutLM}^{\dagger}_{\text{BASE}}$     & 0.6562 & 0.7456 & 0.6980 \\ \midrule
\multirow{4}{*}{11M} & \multirow{2}{*}{1}    & $\text{LayoutLM}_{\text{BASE}}$ \citep{LayoutLM} & 0.7464 & 0.7815 & 0.7636 \\
                     &   & $\text{LayoutLM}^{\dagger}_{\text{BASE}}$     & 0.7384 & 0.8022 & 0.7689 \\ \cmidrule{2-6}
                     & \multirow{2}{*}{2}    & $\text{LayoutLM}_{\text{BASE}}$ \citep{LayoutLM} & 0.7597 & 0.8155 & 0.7866 \\
                     &   & $\text{LayoutLM}^{\dagger}_{\text{BASE}}$     & 0.7612 & 0.8188 & 0.7889 \\ \bottomrule
\end{tabular}
\caption{Sanity checking of LayoutLM$^\dagger$ by comparing its performances on FUNSD EE task from the reported scores in \citet{LayoutLM}. LayoutLM$^\dagger$ is trained in our experimental settings including pre-training datasets and hardware devices.}
\label{tbl:reproduce_layoutlm}
\vspace{-0.25em}
\end{table}

\begin{table}[t]
\centering
\begin{tabular}{l|rrr|rrr}
 & \multicolumn{3}{c|}{BIO tagger} & \multicolumn{3}{c}{SPADE decoder} \\
\multicolumn{1}{c|}{Model} & \multicolumn{1}{c}{F} & \multicolumn{1}{c}{S} & \multicolumn{1}{c}{C} & \multicolumn{1}{|c}{F} & \multicolumn{1}{c}{S} & \multicolumn{1}{c}{C} \\

\toprule
BERT$_{\text{BASE}}$            & 60.92 & 93.67 & 93.13 & 63.38 & 93.09 & 95.16 \\
LayoutLM$^*_{\text{BASE}}$      & 78.54 & 95.11 & 96.26 & 78.47 & 93.33 & 96.71 \\
LayoutLMv2$^*_{\text{BASE}}$    & 81.89 & 96.09 & 96.05 & 78.16 & 95.32 & 96.13 \\
BROS$_{\text{BASE}}$            & \textbf{83.05} & \textbf{96.28} & \textbf{96.50} & \textbf{81.61} & \textbf{95.70} & \textbf{96.73} \\
\midrule
\midrule
BERT$_{\text{LARGE}}$           & 64.17 & 94.25 & 94.74 & 65.23 & 93.40 & 95.30 \\
LayoutLM$^*_{\text{LARGE}}$     & 79.27 & 95.36 & 96.12 & 48.30 & 93.90 & 95.63 \\
LayoutLMv2$^*_{\text{LARGE}}$   & 83.59 & 96.39 & 97.24 & 83.00 & 96.62 & 97.18 \\
BROS$_{\text{LARGE}}$           & \textbf{84.52} & \textbf{96.62} & \textbf{97.28} & \textbf{83.23} & \textbf{96.69} & \textbf{97.32} \\
\bottomrule

\end{tabular}
\caption{Performance comparisons of BIO tagger and SPADE decoder for the three EE tasks \underline{\textit{with}} the order information of text blocks.}
\label{tbl:compare_bio_spade_w_order_ee}
\vspace{-0.5em}
\end{table}

\begin{table}[h]
\centering
\begin{tabular}{c|c|c}

Model & \multicolumn{1}{c|}{Speed (BASE)} & \multicolumn{1}{c}{Speed (LARGE)} \\
\toprule
LayoutLM$^*$     & 0.0413s  {\scriptsize $\pm$ 0.0008}  & 0.1182s {\scriptsize $\pm$ 0.0027} \\
LayoutLMv2$^*$   & 0.1438s  {\scriptsize $\pm$ 0.0009}  & 0.2330s {\scriptsize $\pm$ 0.0011} \\
BROS         & 0.0721s  {\scriptsize $\pm$ 0.0007}  & 0.1794s {\scriptsize $\pm$ 0.0021} \\
\bottomrule
\end{tabular}
\caption{Inference speed (Avg. of 50 samples on T4 GPU).}
\label{tbl:infer_speed}
\vspace{-0.5em}
\end{table}

\section{Compare Published LayoutLM Model and Our Own Implementation}

Table~\ref{tbl:reproduce_layoutlm} compares our implementation of LayoutLM from the reported scores in \citet{LayoutLM}.
As can be seen, multiple experiments are conducted according to the number of pre-training data.
Our implementation, referred to LayoutLM$^\dagger$, shows comparable performances over all settings.

\section{Compare BIO Tagger and SPADE Decoder}

Table~\ref{tbl:compare_bio_spade_w_order_ee} shows the performance comparisons of BIO tagger and SPADE decoder for the three EE tasks when the order of text blocks is given.
There is not much difference in performance between the BIO tagger and the SPADE decoder in this case.
Interestingly, the performance of BERT on FUNSD and CORD is higher when using SPADE decoder than when using BIO tagger.
This seems to be because 2D layout information can be utilized by using the SPADE decoder.
In FUNSD EE task, the performance of LayoutLM$^*_{\text{LARGE}}$ is unstable to have a standard deviation of 12.51.
BROS shows the best performance in all tasks regardless of the parsers.

\section{Compare the Inference Speed of the Models}

Table~\ref{tbl:infer_speed} shows the inference speed of the models.
Since BROS considers relative positions for all text block pairs, it is slower than LayoutLM, but faster than LayoutLMv2 using image features.
It should be noted that BROS shows the best performance among them.

\clearpage

\begin{figure}[t]
    \centering
    \includegraphics[width=0.7\linewidth]{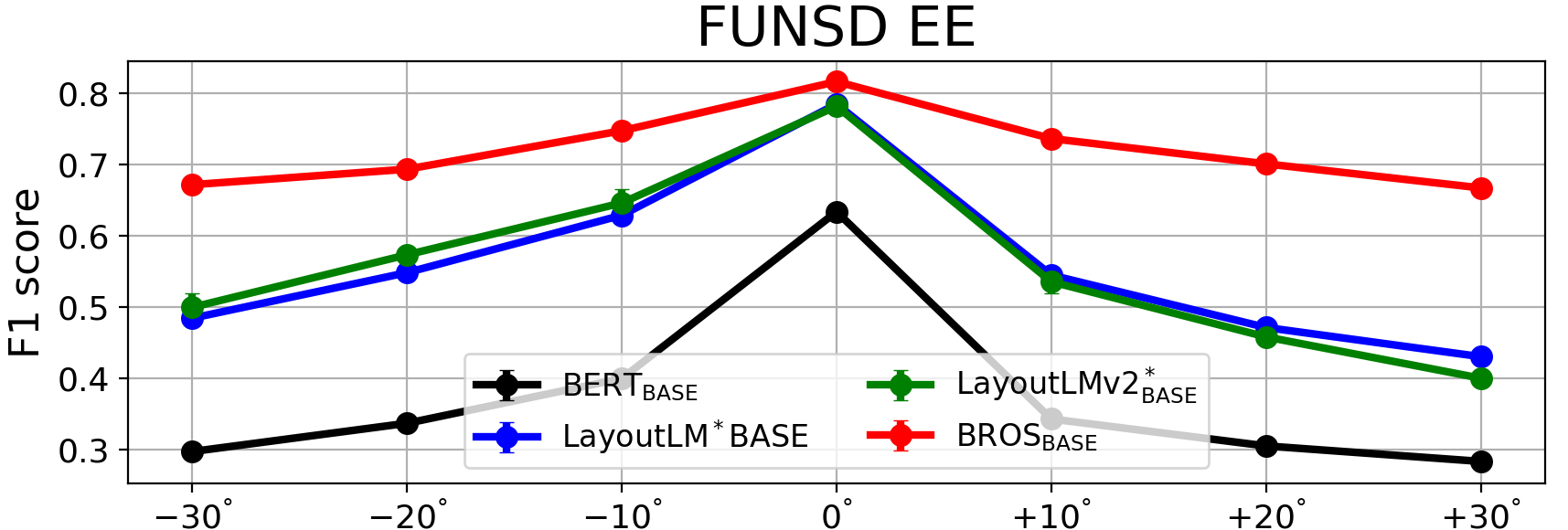}
    \caption{Results of rotated FUNSD EE task.}
    \label{fig:rotated_funsd_appendix}
    \vspace{-0.5em}
\end{figure}

\section{Experiments on Rotated Images}

To test in another imprecise text serialization setup, we conducted experiment on the rotated images in which the image and locations of all text blocks are rotated and the block order is re-serialized (yx-).
Figure~\ref{fig:rotated_funsd_appendix} shows the results with the same trend as Table~\ref{tbl:tbl_sorting_funsd_ee}.

\end{document}